\documentclass{article}
\usepackage{proceed2e}

\usepackage[T1]{fontenc}

\usepackage{amsmath}
\usepackage{amssymb}
\usepackage{amsthm}
\usepackage{float}
\usepackage[colorlinks=true, urlcolor=blue, citecolor=blue]{hyperref}
\usepackage{mathtools}
\usepackage{multicol}
\usepackage{nicefrac}
\usepackage{underscore}
\usepackage[dvipsnames]{xcolor}

\usepackage{tikz}
\usetikzlibrary{decorations.markings, fadings, shapes}
\definecolor{myblue}{RGB}{16,129,216} 
\definecolor{myred}{RGB}{219,4,27} 
\tikzfading[name = fade out, left color=transparent!75, right color=transparent!20, shading angle = 30]
\tikzstyle{n} = [draw, circle, thick, minimum size = 3.5em, node distance = 4.5em, xshift = 1.5em, scale = .75]
\tikzstyle{w} = [n, preaction = {fill = gray, path fading = fade out}]
\tikzstyle{o} = [n, preaction = {fill = myblue, path fading = fade out}]
\tikzstyle{m} = [n, preaction = {left color = myblue, right color = myred, path fading = fade out}]
\tikzstyle{a} = [n, preaction = {fill = myred, path fading = fade out}]
\tikzstyle{e} = [->, thick]
\tikzstyle{d} = [e, dashed]
\tikzstyle{de} = [e, <->]

\newcommand\numberthis{\addtocounter{equation}{1}\tag{\theequation}}

\newtheorem{defn}{Definition}
\DeclareMathOperator{\DKL}{D_{KL}}
\DeclareMathOperator{\E}{E}
\DeclareMathOperator{\I}{I}
\newcommand{\C}[1]{\mathcal{#1}}
\newcommand{\eqn}{\begin{eqnarray*}}
\newcommand{\enn}{\end{eqnarray*}}

\begin{document}

\title{Optimal Selective Attention in Reactive Agents}
\subtitle{Technical Report}
\author{ {\bf Roy Fox \qquad\qquad\qquad Naftali Tishby} \\\\
School of Computer Science and Engineering \\
The Hebrew University}
\date{}

\maketitle

\begin{abstract}
In POMDPs, information about the hidden state, delivered through observations, is both valuable to the agent, allowing it to base its actions on better informed internal states, and a "curse", exploding the size and diversity of the internal state space. One attempt to deal with this is to focus on reactive policies, that only base their actions on the most recent observation. However, even reactive policies can be demanding on resources, and agents need to pay selective attention to only some of the information available to them in observations. In this report we present the minimum-information principle for selective attention in reactive agents. We further motivate this approach by reducing the general problem of optimal control in POMDPs, to reactive control with complex observations. Lastly, we explore a newly discovered phenomenon of this optimization process --- period doubling bifurcations. This necessitates periodic policies, and raises many more questions regarding stability, periodicity and chaos in optimal control.
\end{abstract}

\section{Introduction}

For an intelligent agent interacting with its environment, information is valuable.
By observing and retaining information about its environment, the agent can form beliefs and make predictions.
It represents these beliefs in an internal state, on which it can then base its actions.

If information about some event in the world is unavailable to the agent, through the lack of observability or attention, its internal state is independent of that event, and so are its actions, potentially incurring otherwise avoidable costs.
The same is true if the information is only partially available, limiting the extent to which the agent's actions can depend on the state of the world.

However, information is also a "curse".
Retaining much information about the world requires the agent to have a large and rich internal state space, representing diverse beliefs.
This leads to complex policies for inference and control, which are computationally hard both to find and to apply.
Designed agents should not be --- and evolved agents are unlikely to be --- more complex than is sufficient for them to perform well.

The "curse of dimensionality" \cite{bellman1957dynamic} is the challenge of representing in the internal state space the entire belief space --- the space of probability distributions over world states.
The volume of this simplex is exponential in the number of world states, and approximate methods \cite{shani2013survey} \cite{roy2005finding} \cite{aberdeen2003revised} \cite{murphy2000survey} are required to explore and represent policies over this space.

The "curse of history" \cite{pineau2003point} results from representing only reachable Bayesian beliefs --- posteriors of the world state given each possible observable history.
The Bayesian belief is a sufficient statistic of the observable history for the world state, keeping all available information about it.
Unfortunately, the size of this space can be exponential in the length of the history.

This realization immediately suggests the idea of truncating the observable history by forgetting older observations.
Taken to the extreme, this leads to reactive agents \cite{littman1994memoryless} \cite{singh1994learning}, in which each internal state can only take into account the most recent observation, discarding the previous internal state.
The internal state space of reactive agents needs not be larger than the observation space, which removes the curse of history in domains where the set of observations is not too large.

\begin{defn}
A \emph{reactive agent} bases its actions only on the most recent observation.
In contrast, a \emph{retentive agent} can base its actions on a memory state, which is updated with each observation, and thus summarizes the entire observable history.
\end{defn}

A drawback of this approach is that, since the history is no longer grounded in a known initial belief, a new challenge arises of identifying which beliefs these internal states represent.
This challenge generally requires forward-backward algorithms \cite{fox2012bounded}, as opposed to fully observable Markov Decision Processes which are solvable by backward (dynamic programming) algorithms \cite{bellman1957dynamic}.

In addition, the original difficulty remains in domains where the observation space is still too large, such as the one presented in Section \ref{sec:reduction}.
In this sense, the curse of history is a special case of the following principle, which we might call the "curse of information".

An agent's input --- its sensors, and its memory when available --- usually contains too much information for the agent to process.
For the agent to encode all of this information in its new internal state, an internal state space is required that is too large to be manageable and utilized by feasible policies.
As a matter of practicality, an agent must have selective attention.
A retentive agent must also have selective retention \cite{fox2012bounded}, which is beyond the scope of this paper.

\begin{defn}
A reactive agent (similarly, a retentive agent), is said to have \emph{selective attention} (resp. \emph{selective retention}) if its internal state has less information about the world state than its observation (resp. observable history) does.
\end{defn}

Reactive policies have been explored before in \cite{littman1994memoryless}, with some of their challenges noted in \cite{singh1994learning}.
A policy-gradient algorithm for finding such policies was presented in \cite{jaakkola1995reinforcement}, which has the nice property of avoiding the forward-backward coupling.
However, the local optimum it finds is not guaranteed to be a fixed point of the value recursion.

Information considerations in dynamical systems were presented in \cite{tishby2011information}.
Algorithms were later introduced for trading off value and information in fully observable Markov Decision Processes \cite{rubin2012trading} and in partially observable ones where actions have no external effect \cite{fox2012bounded}.

This paper offers three novel contributions, in each of the following sections.

Section \ref{sec:reduction} shows that reactive policies are as expressive as retentive policies, under proper redefinition of the model.
This motivates our focus on reactive agents, at the same time that it demands a more principled cure for the curse of information than simply discarding the memory.

Section \ref{sec:information} provides such a principle, namely the minimum-information principle.
We present the principle and formalize it, discuss its relation to source coding, and give an algorithm for its numeric solution.

Section \ref{sec:periodic} demonstrates a newly discovered phenomenon in optimal control, namely the occurrence of bifurcations when attention is traded off with external cost.
This is illustrated using two examples.

We conclude with a short discussion of these contributions and their consequences.

\section{Preliminaries}

We model the interaction of an intelligent agent with its environment using the formalism of Partially Observable Markov Decision Processes (POMDPs).
A POMDP is a discrete-time dynamical system with state $s_t\in\C S$.
In time $t$, the system emits an observation $o_t\in\C O$ with probability $\sigma(o_t|s_t)$.
It then receives from the interacting agent an input action $a_t\in\C A$, and transitions to a new state $s_{t+1}$ with probability $p(s_{t+1}|s_t,a_t)$.
For our purposes here, the sets $\C S$, $\C O$ and $\C A$ are finite, and we are only concerned with stationary (time-invariant) POMDPs, where the model parameters $p$ and $\sigma$ are fixed for every time step.

\begin{figure}
\center
\begin{tikzpicture}
\node (s)		[w, minimum size=5em]						{world};
\node (o)		[w, minimum size=5em, below of=s, xshift=-7em, yshift=-2em]	{sensor};
\node (a)		[w, minimum size=5em, below of=s, xshift=3.2em, yshift=-2em]		{actuator};
\draw (s) -- (o)	[e] node [midway, left, yshift=1em, scale=.7] {\begin{tabular}{c}observation\\$\sigma(o|s)$\end{tabular}};
\draw (o) -- (a)	[e] node [midway, below, scale=.7] {\begin{tabular}{c}policy\\$\pi(a|o)$\end{tabular}};
\draw (a) -- (s)	[e] node [midway, right, yshift=1em, scale=.7] {\begin{tabular}{c}action\\$p(s'|s,a)$\end{tabular}};
\end{tikzpicture}
\caption{Schematic model of a reactive agent interacting with its environment}
\label{fig:agent}
\end{figure}
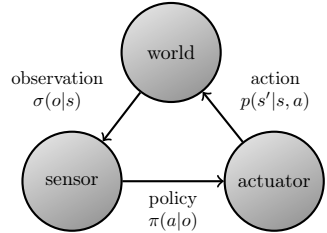

A reactive agent has no internal memory state, and can only base its actions on the most recent observation.
The agent consists of two modules, the sensor making the observation $o_t$ and the actuator taking the action $a_t$ (Figure \ref{fig:agent}).
The reactive policy $\pi$ of the agent is implemented by linking the two modules through a communication channel, such that the action $a_t$ is taken with probability $\pi_t(a_t|o_t)$ in reaction to observation $o_t$.
The policy is called periodic with period $\C T$ if $\pi_t=\pi_{t+\C T}$ for every time step $t$.
The policy is called stationary if it has period 1, i.e. $\pi_t$ is fixed for every time step.

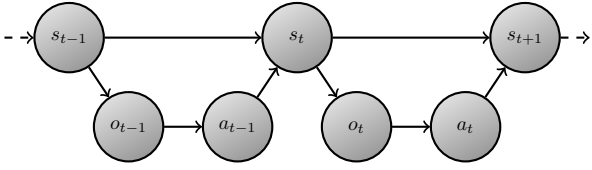
\begin{figure}
\center
\begin{tikzpicture}
\node (sp)		[w]						{$s_{t-1}$};
\node (op)		[w, below of=sp, xshift=1em]	{$o_{t-1}$};
\node (ap)		[w, right of=op, xshift=-1em]				{$a_{t-1}$};
\node (st)		[w, right of=sp, xshift=5em]	{$s_t$};
\node (ot)		[w, below of=st, xshift=1em]				{$o_t$};
\node (at)		[w, right of=ot, xshift=-1em]				{$a_t$};
\node (sf)		[w, right of=st, xshift=5em]		{$s_{t+1}$};
\coordinate	[left of=sp, xshift=.4em]		(sh);
\coordinate	[right of=sf, xshift=-.4em]		(sn);
\draw (sh) -- (sp)	[d];
\draw (sp) -- (op)	[e, label={[right]{$\sigma$}}];
\draw (op) -- (ap)	[e, label={[below]{$\pi$}}];
\draw (sp) -- (st)	[e];
\draw (ap) -- (st)	[e, label={[above left]{$p$}}];
\draw (st) -- (ot)		[e, label={[right]{$\sigma$}}];
\draw (ot) -- (at)		[e, label={[below]{$\pi$}}];
\draw (st) -- (sf)		[e];
\draw (at) -- (sf)		[e, label={[above left]{$p$}}];
\draw (sf) -- (sn)	[d];
\end{tikzpicture}
\caption{Graphical model of a reactive agent interacting with its environment}
\label{fig:process}
\end{figure}

The model and the policy together induce a stochastic process over the variables $\{s_t,o_t,a_t\}$ (Figure \ref{fig:process}).
Due to the agent's lack of memory, the states $\{s_t\}$ form a Markov chain.
In the following we always assume that the process is ergodic.
This implies that, if the agent policy has period $\C T$, then for each phase $0\le t<\C T$ there exists a unique marginal distribution $\bar p_t(s_t)$ that is a fixed point of the $\C T$-step forward recursion
$$\bar p_t(s_{t+\C T})=\sum_{s_t}\bar p_t(s_t)P_{t,\pi}(s_{t+\C T}|s_t)$$
with
$$P_{t,\pi}(s_{t+\C T}|s_t)=\sum_{s_{t+1},\ldots,s_{t+\C T-1}}\prod_{\tau=t}^{t+\C T-1}P_{\pi_\tau}(s_{\tau+1}|s_\tau)$$
and
$$P_{\pi_\tau}(s_{\tau+1}|s_\tau)=\sum_{o_\tau,a_\tau}\sigma(o_\tau|s_\tau)\pi_\tau(a_\tau|o_\tau)p(s_{\tau+1}|s_\tau,a_\tau).$$
These marginal distributions are therefore periodic with the same period $\C T$, i.e. $\bar p_t=\bar p_{t+\C T}$, and inside a cycle the phases are linked through the 1-step forward recursion
\begin{equation}
\label{eq:forward}
\bar p_{t+1}(s_{t+1})=\sum_{s_t}\bar p_t(s_t)P_{\pi_t}(s_{t+1}|s_t).
\end{equation}

The marginal distributions also induce beliefs
$$b_t(s_t|o_t)=\frac{\bar p_t(s_t)\sigma(o_t|s_t)}{\bar\sigma_t(o_t)},$$
with
$$\bar\sigma_t(o_t)=\sum_{s_t}\bar p_t(s_t)\sigma(o_t|s_t).$$
The belief is the posterior distribution of the state given the observation.

In this paper we will have the agent incur an external nominal cost $c(s_t,a_t)$ when it takes action $a_t$ in state $s_t$, and measure the quality of a policy by the long-term average expected cost
$$\C C=\lim_{T\to\infty}\frac1T\sum_{t=0}^{T-1}\E[c(s_t,a_t)]$$
in the stochastic process that the policy induces.
If the policy has period $\C T$ and the process is at its periodic marginal distribution, then
$$\C C=\frac1{\C T}\sum_{t=0}^{\C T-1}\sum_{s_t,o_t,a_t}\bar p_t(s_t)\sigma(o_t|s_t)\pi_t(a_t|o_t)c(s_t,a_t).$$

This undiscounted expected cost is appropriate for studying stationary processes.
In contrast, discounting the cost by $\gamma^t$ emphasizes transient effects, up to horizon $O(\tfrac1{1-\gamma})$.
A related fault with discounting in reactive policies is discussed in~\cite{singh1994learning}.

\section{Reduction from retentive to reactive policies}
\label{sec:reduction}

\begin{figure}
\center
\begin{tikzpicture}
\node (sp)		[w]						{$s_{t-1}$};
\node (op)		[w, below of=sp, xshift=1em]	{$o_{t-1}$};
\node (mp)	[w, below of=op, xshift=1em]	{$m_{t-1}$};
\node (ap)		[w, above of=mp, xshift=1em]				{$a_{t-1}$};
\node (st)		[w, above of=ap, xshift=1em]	{$s_t$};
\node (ot)		[w, below of=st, xshift=1em]				{$o_t$};
\node (mt)		[w, below of=ot, xshift=1em]	{$m_t$};
\node (at)		[w, above of=mt, xshift=1em]				{$a_t$};
\node (sf)		[w, above of=at, xshift=1em]		{$s_{t+1}$};
\coordinate	[left of=sp, xshift=.4em]		(sh);
\coordinate	[right of=sf, xshift=-.4em]		(sn);
\coordinate	[left of=mp, xshift=-2em]		(mh);
\coordinate	[right of=mt, xshift=2em]		(mf);
\draw (sh) -- (sp)	[d];
\draw (sp) -- (op)	[e, label={[right]{$\sigma$}}];
\draw (mh) -- (mp)	[d];
\draw (op) -- (mp)	[e, label={[below left]{$q$}}];
\draw (mp) -- (ap)	[e, label={[right]{$\pi$}}];
\draw (sp) -- (st)	[e];
\draw (ap) -- (st)	[e, label={[above left]{$p$}}];
\draw (st) -- (ot)		[e, label={[right]{$\sigma$}}];
\draw (mp) -- (mt)	[e];
\draw (ot) -- (mt)	[e, label={[below left]{$q$}}];
\draw (mt) -- (at)	[e, label={[right]{$\pi$}}];
\draw (st) -- (sf)		[e];
\draw (at) -- (sf)		[e, label={[above left]{$p$}}];
\draw (mt) -- (mf)	[d];
\draw (sf) -- (sn)	[d];
\end{tikzpicture}
\caption{Graphical model of a retentive agent interacting with its environment}
\label{fig:retentive}
\end{figure}
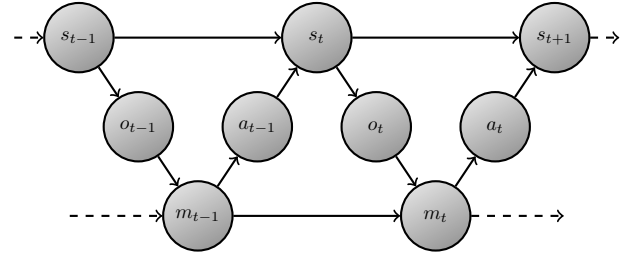

Consider a retentive agent \cite{aberdeen2003policy} \cite{fox2012bounded} interacting with a POMDP (Figure \ref{fig:retentive}).
The agent has an internal state $m_t\in\C M$, and an inference policy $q_t$ controlling it, such that with probability $q_t(m_t|m_{t-1},o_t)$ the memory state $m_{t-1}$ is updated to $m_t$ upon observing $o_t$ in time step $t$.
The control policy $\pi_t(a_t|m_t)$ is allowed to depend not only on the most recent observation, but on the summary of the entire observable history represented in $m_t$.

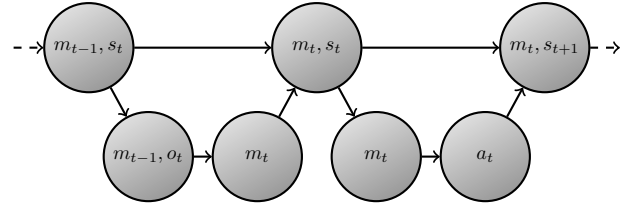
\begin{figure}
\center
\begin{tikzpicture}
\node (sp)		[w, minimum size=4.5em]						{$m_{t-1},s_t$};
\node (op)		[w, below of=sp, xshift=1em, yshift=-1em, minimum size=4.5em]	{$m_{t-1},o_t$};
\node (ap)		[w, right of=op, xshift=-1em, minimum size=4.5em]								{$m_t$};
\node (st)		[w, right of=sp, xshift=5em, minimum size=4.5em]	{$m_t,s_t$};
\node (ot)		[w, below of=st, xshift=1em, yshift=-1em, minimum size=4.5em]								{$m_t$};
\node (at)		[w, right of=ot, xshift=-1em, minimum size=4.5em]								{$a_t$};
\node (sf)		[w, right of=st, xshift=5em, minimum size=4.5em]	{$m_t,s_{t+1}$};
\coordinate	[left of=sp, xshift=0em]		(sh);
\coordinate	[right of=sf, xshift=0em]		(sn);
\draw (sh) -- (sp)	[d];
\draw (sp) -- (op)	[e, label={[right]{$\sigma$}}];
\draw (op) -- (ap)	[e, label={[below]{$q$}}];
\draw (sp) -- (st)	[e];
\draw (ap) -- (st)	[e];
\draw (st) -- (ot)		[e];
\draw (ot) -- (at)		[e, label={[below]{$\pi$}}];
\draw (st) -- (sf)		[e];
\draw (at) -- (sf)		[e, label={[above left]{$p$}}];
\draw (sf) -- (sn)	[d];
\end{tikzpicture}
\caption{Reduction from a retentive policy to a reactive policy}
\label{fig:reduction}
\end{figure}

In a given POMDP, retentive policies $(q,\pi)$ are much more expressive and powerful than reactive policies.
Interestingly, however, there exists another (time-variant) POMDP in which $\pi'=(q,\pi)$ can be implemented as a reactive policy (Figure \ref{fig:reduction}).
This new POMDP is similar in spirit to the cross-product MDP~\cite{meuleau1999solving}, and the distinction between them is discussed below.

Formally, let the state space of the new POMDP be $\C S'=\C M\times\C S$, the observation space $\C O'=\C M\times(\C O\cup\{\bot\})$, where $\bot$ is some null-observation symbol, and the action space $\C A'=\C M\cup\C A$.
Let the dynamics advance at twice the frequency, with each time step taking half as long.
The state at time $t$ is $s'_t=(m_{t-1},s_t)$, and it emits an observation with distribution
$$\sigma'_t((m_{t-1},o_t)|(m_{t-1},s_t))=\sigma(o_t|s_t).$$
The agent, upon observing $(m_{t-1},o_t)$, can apply its inference policy to generate the next memory state $m_t$.
It then takes the "action" of committing $m_t$ to "external storage"
$$p'_t((m_t,s_t)|(m_{t-1},s_t),m_t)=1.$$
In this new state at time $t+\tfrac12$, the committed memory state is observable
$$\sigma'_{t+\tfrac12}((m_t,\bot)|(m_t,s_t))=1,$$
and the agent can apply its control policy to take the action $a_t$, thus controlling the transition
$$p'_{t+\tfrac12}((m_t,s_{t+1})|(m_t,s_t),a_t)=p(s_{t+1}|s_t,a_t).$$

Note that it should be inadmissible for the agent to commit a memory state in a non-integer time step, or take an action in an integer one.
This can be enforced by penalizing the wrong type of action, which is the main reason that the new POMDP needs to be time-variant.

Assuming that the agent follows these restrictions, the new POMDP induces the same stochastic process over the variables $\{s_t,o_t,m_t,a_t\}$ as the original one for any given policy, establishing the reduction.

Our reduction is related to the cross-product MDP of~\cite{meuleau1999solving}.
However, the two models have different formulations that serve their different purposes --- where the cross-product MDP creates structure to be exploited in planning algorithms, our formulation flattens this structure to reduce the problem to a simpler one.
To achieve this, instead of the implicit restriction in~\cite{meuleau1999solving} that policies depend only on the agent state, we model the same constraint explicitly as partial observability.
Furthermore, by breaking each time step into two we avoid the exponential action space of the cross-product MDP.

Lastly, an important issue to consider is the memory state space $\C M$.
The standard approach in the reinforcement learning literature is to have $\C M$ be the belief space, the simplex of distributions over $\C S$, and $q$ the Bayesian inference policy\footnote{It is also common to have actions as part of the observable history, which our notation allows but does not require.}.
Such a choice would make $\C S'$, $\C O'$ and $\C A'$ uncountable, as opposed to our usual assumption that these sets are finite.

Alternatively, we can have in $\C M$ only reachable beliefs.
If the support of the inference policy $q$ is finite\footnote{For example, the Bayesian inference policy is deterministic.}, then over a finite horizon only a finite number of beliefs are reachable.
Unfortunately, due to the "curse of history", this number is exponential in the horizon, which renders this reduction --- and indeed many existing approaches to POMDPs --- impractical.

This difficulty underlines the need for selective attention.
Theoretically, the support of $m_t$ needs never be larger than that of $(m_{t-1},o_t)$, at least in terms of sufficient inference.
However, it should practically be much smaller than that --- roughly the same size as the support of $m_{t-1}$ --- if the agent is to interact with the system for significant horizons without exploding in complexity.
The ability of the agent to selectively attend to its input, whether from sensors or from memory, and to retain not all, but only the most useful information, is key to reducing this complexity.

This is the approach taken by Finite State Controllers (FSCs)~\cite{poupart2003bounded}, where the number of memory states is fixed.
Several heuristic algorithms exist for finding a good FSC, however this problem is hard and highly non-convex.
The policy of a FSC is time-invariant, and as we see in Section~\ref{sec:periodic} a stationary Bellman-optimal solution is generally not stable.

\section{Minimum-information principle}
\label{sec:information}

Our guiding principle in formalizing selective attention is the reduction of information complexity, as measured by the Shannon mutual information between the observation $o_t$ and the action $a_t$.
We first present the principle, and then justify it by relating it to source coding.
We note that numerous other justifications and connections exist, some discussed previously \cite{kappen2012optimal} \cite{tishby2011information} \cite{todorov2006linearly} \cite{rose1998deterministic} \cite{jaynes2003probability}, and some should be explored further, particularly in the context of POMDP planning.

The pointwise mutual information between $o_t$ and $a_t$ in time step $t$ is given by
$$i_t(o_t,a_t)=\log\frac{\pi_t(a_t|o_t)}{\bar\pi_t(a_t)},$$
with
\begin{equation}
\label{eq:margin}
\bar\pi_t(a_t)=\sum_{o_t}\bar\sigma_t(o_t)\pi_t(a_t|o_t).
\end{equation}
This can be thought of as the internal informational cost of choosing action $a_t$ in reaction to observation $o_t$.
The long-term average expectation of this internal cost, similar to the external cost, is
$$\C I=\lim_{T\to\infty}\frac1T\sum_{t=0}^{T-1}\E[i_t(o_t,a_t)].$$
If the policy has period $\C T$ and the process is at its periodic marginal distribution then
\begin{align*}
\C I=&\frac1{\C T}\sum_{t=0}^{\C T-1}\sum_{o_t,a_t}\bar\sigma_t(o_t)\pi_t(a_t|o_t)i_t(o_t,a_t)\\
=&\frac1{\C T}\sum_{t=0}^{\C T-1}\DKL[\pi_t\|\bar\pi_t]=\frac1{\C T}\sum_{t=0}^{\C T-1}\I[o_t;a_t].
\end{align*}
Here $\DKL[\pi_t\|\bar\pi_t]$ is the Kullback-Leibler divergence of $\pi_t$ from $\bar\pi_t$, and $\I[o_t;a_t]$ is the Shannon mutual information between $o_t$ and $a_t$.

$\DKL[\pi_t\|\bar\pi_t]$ is a measure of the cognitive effort required for the agent to diverge from a passive, uncontrolled policy $\bar\pi_t$ to an active, controlled policy $\pi_t$.
Unlike \cite{kappen2012optimal} \cite{tishby2011information} \cite{todorov2006linearly}, we allow the passive policy, as well as the active one, to be designed or evolved.
The uncontrolled policy that minimizes the informational cost $\C I$ is the appropriate marginal distribution of the action \eqref{eq:margin} \cite{cover2012elements}.

Among agents incurring external cost $\C C\le C$, the simplest agent, in some sense, minimizes the internal cost $\C I$.
In other words, the agent needs to trade off its external and internal costs.
To link these views, the Lagrange multiplier $\beta$ corresponding to the constraint $\C C\le C$ in the optimization of $\C I$ is a conversion rate between the two types of cost.
We can then write the total cost as
$$\C F=\tfrac1\beta\C I+\C C.$$
$\C F$ is called the free energy, due to its similarity to the quantity of the same name in statistical physics, with $\beta$ taking the part of the inverse temperature.

For a given $\beta$, the agent chooses its policy so as to minimize the free energy, under two constraints.
First, the dynamics of the system follow the forward recursion \eqref{eq:forward}.
Second, $\bar p_t$, $\pi_t(\cdot|o_t)$ and $\bar\pi_t$ need to be probability distributions, each summing to 1.
The constraints that they are non-negative can be ignored, since they will be either inactive or weakly active.

This gives for horizon $T$ the Lagrangian $\C L_{\bar p,\pi,\bar\pi}$
$$=\frac1T\sum_{t=0}^{T-1}\Bigg(\sum_{s_t,o_t,a_t}\bar p_t(s_t)\sigma(o_t|s_t)\pi_t(a_t|o_t)f_t(s_t,o_t,a_t)$$
$$+\sum_{s_{t+1}}\nu_{t+1}(s_{t+1})\Bigg(\sum_{s_t}\bar p_t(s_t)P_{\pi_t}(s_{t+1}|s_t)-\bar p_{t+1}(s_{t+1})\Bigg)$$
$$-\varphi_t\Bigg(\sum_{s_t}\bar p_t(s_t)-1\Bigg)+\eta_t\Bigg(\sum_{a_t}\bar\pi_t(a_t)-1\Bigg)$$
$$+\sum_{o_t}\lambda_t(o_t)\Bigg(\sum_{a_t}\pi_t(a_t|o_t)-1\Bigg)\Bigg),$$
with
$$f_t(s_t,o_t,a_t)=\tfrac1\beta i_t(o_t,a_t)+c(s_t,a_t).$$

\subsection{Necessary conditions for optimality}

This optimization problem is far from convex, and no efficient algorithm is known for finding the global optimum.
Indeed, as $\beta$ tends to infinity, the agent's policy becomes deterministic, and some problems involving deterministic reactive policies are known to be NP-complete \cite{littman1994memoryless}.

Nevertheless, we can consider local minima by finding the first-order necessary conditions for a solution to be optimal.
That is, we differentiate the Lagrangian by each of its parameters, and require that this derivative equals 0.

For $\bar p$, this gives us a backward recursion
\begin{align*}
\nu_t(s_t)=&\sum_{o_t,a_t}\sigma(o_t|s_t)\pi_t(a_t|o_t)f_t(s_t,o_t,a_t)\\
+&\sum_{s_{t+1}}P_{\pi_t}(s_{t+1}|s_t)\nu_{t+1}(s_{t+1})-\varphi_t.
\numberthis\label{eq:backward}
\end{align*}
Due to overconstraining, we have some degrees of freedom in choosing the multipliers to satisfy the Karush-Kuhn-Tucker conditions \cite{boyd2004convex}.
If the policy has period $\C T$, we will choose $\varphi_t$ to also have period $\C T$ and satisfy
$$\frac1{\C T}\sum_{t=0}^{\C T-1}\varphi_t=\C F,$$
so that $\nu_t$ also has period $\C T$.
Thus $\nu_t(s_t)$ measures the fluctuation from the average free energy $\C F$ of the state $s_t$ in phase $t$ of the cycle.

The first-order necessary conditions for $\pi$ are
\begin{equation}
\label{eq:opt}
\pi_t(a_t|o_t)=\frac1{Z_t(o_t)}\bar\pi_t(a_t)\exp(-\beta d_t(o_t,a_t))
\end{equation}
with
\begin{align*}
d_t(o_t,a_t)=&\sum_{s_t}b_t(s_t|o_t)c(s_t,a_t)\\
+&\sum_{\mathclap{s_t,s_{t+1}}}b_t(s_t|o_t)p(s_{t+1}|s_t,a_t)\nu_{t+1}(s_{t+1})
\end{align*}
and the normalizing partition function
\eqn
Z_t(o_t)=\sum_{a_t}\bar\pi_t(a_t)\exp(-\beta d_t(o_t,a_t)),
\enn
and for $\bar\pi$ we have \eqref{eq:margin} as promised.

As $\beta$ tends to infinity, the optimal policy in \eqref{eq:opt} becomes deterministic.
Together with \eqref{eq:backward}, it becomes a Bellman equation \cite{bellman1957dynamic}.

For finite $\beta$, on the other hand, the optimal policy is stochastic, which is a welcome outcome in many respects.
The best deterministic reactive policy is generally arbitrarily worse than the optimal stochastic reactive policy \cite{singh1994learning}.
Optimality in reactive policies requires stochasticity.
Unfortunately, many planning algorithms rely on the smaller space of deterministic policies (e.g. \cite{pineau2003point}), and others lack a principle by which to gauge the optimal amount of uncertainty in the agent's actions (e.g. \cite{aberdeen2003policy}).
We propose minimum information as such a principle.

Furthermore, in practice the model used for planning is itself uncertain.
Using deterministic policies could overfit to the available data, and hinder further learning \cite{sutton1998introduction}.
Information considerations provide a principled way of fitting the uncertainty of the policy to the uncertainty of the model \cite{rubin2012trading}.

In reinforcement learning, it is common to use soft-max to obtain stochastic planning and exploration policies \cite{aberdeen2003policy} \cite{sutton1998introduction}.
Note that soft-max is a special case of \eqref{eq:opt}, with the uniform prior instead of the marginal $\bar\pi_t$.
That information theory provides a better principle for stochasticity in optimization is further illustrated in the next subsection.

\subsection{Sequential rate-distortion}

The form of \eqref{eq:opt} and \eqref{eq:margin} may be familiar as the solution to the rate-distortion problem of lossy source coding \cite{cover2012elements}.
Indeed, minimum-information optimal control can be construed as a sequential rate-distortion problem \cite{tatikonda1998control}.

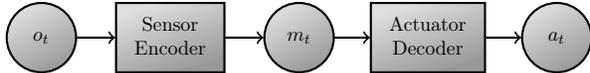
\begin{figure}
\center
\begin{tikzpicture}
\node (o)		[w]					{$o_t$};
\node (enc)	[w, rectangle, minimum width=5em, right of=o]	{\begin{tabular}{c}Sensor\\Encoder\end{tabular}};
\node (m)		[w, right of=enc]			{$m_t$};
\node (dec)	[w, rectangle, minimum width=5em, right of=m]	{\begin{tabular}{c}Actuator\\Decoder\end{tabular}};
\node (a)		[w, right of=dec]			{$a_t$};
\draw (o) -- (enc)	[e];
\draw (enc) -- (m)	[e];
\draw (m) -- (dec)	[e];
\draw (dec) -- (a)	[e];
\end{tikzpicture}
\caption{Reactive policy as a source-coding problem}
\label{fig:channel}
\end{figure}

The reactive agent's policy is a channel from its sensor to its actuator (Figure \ref{fig:channel}).
Following the classic model of such a channel, the sensor can be considered an encoder which, upon observing the "source" $o_t$, chooses a "codeword" $m_t$.
It transmits it to the actuator, a decoder which then "reconstructs" the intended $a_t$.

For a given time step $t$, and with a source distribution $\bar\sigma_t(o_t)$ and a distortion function $d_t(o_t,a_t)$ fixed, this would be a standard source-coding problem.
Let a feasible agent be one achieving at most $D$ expected distortion
$$\sum_{o_t,a_t}\bar\sigma_t(o_t)\pi_t(a_t|o_t)d_t(o_t,a_t)\le D.$$
Now suppose we are interested in the feasible agent with the simplest internal state space, as measured by the size of the "codebook" $\C M$.
The rate-distortion theorem \cite{cover2012elements} states that the simplest feasible agent is the one minimizing $\I[o_t;a_t]$.

In a sequential rate-distortion problem, the solution $\pi_t$ in time step $t$ affects future source distributions $\bar\sigma_\tau$ in time steps $\tau>t$, as well as past distortions $d_\tau$ in time steps $\tau<t$.
This creates a coupling between the forward inference process of computing marginal distributions, and the backward control process of computing value functions.
This is further complicated in partially observable processes, where $d_t$ also depends on the forward inference process through the beliefs $b_t$.

Coupling makes sequential rate-distortion complex, both conceptually and computationally.
Conceptually, the results of rate-distortion theory are no longer known to hold in the sequential case.
If we nevertheless accept the minimum information as a solid guiding principle in our optimization, we find that this optimization is computationally hard.
We can optimize the policy in each time step given the other time steps with algorithms like Blahut-Arimoto \cite{osullivan1998alternating}.
However the forward-backward algorithm for finding the overall policy is only guaranteed to converge to a local optimum \cite{fox2012bounded}.

\subsection{Optimization algorithm}
\label{sec:algorithm}

The forward recursion \eqref{eq:forward}, the backward recursion \eqref{eq:backward}, the optimal policy \eqref{eq:opt} and its marginal \eqref{eq:margin} are necessary conditions for a solution to be optimal.
They also provide an algorithm for finding a good solution: iteratively plug the current solution in the right-hand side of one of the equations, to obtain a better solution, until (asymptotically) no such improvement is possible.
Many existing algorithms employ a similar scheme.
For example, in the Generalized Policy Iteration algorithm for planning in MDPs \cite{sutton1998introduction}, there is some schedule for alternating between\footnote{A forward equation is not needed in fully observable problems if attention is not selective.} policy evaluation, a variant of \eqref{eq:backward}, and policy improvement, a variant of \eqref{eq:opt} with $\beta\to\infty$.

A sophisticated schedule can guarantee that the solution improves monotonically with each iteration \cite{fox2012bounded}.
Here we suggest the following simpler schedule, for which such a guarantee does not hold, but which empirically converges to good solutions in practice.

Repeat until convergence:
\begin{enumerate}
	\item Compute the marginal $\bar\pi$ given the current solution for $\pi$, by applying \eqref{eq:margin}.
	\item Compute the value function $\nu$ given the current solution for $\bar p$, $\pi$ and $\bar\pi$. This can be done by iteratively applying \eqref{eq:backward} until it converges, or by solving it as a system of linear equations.
	\item In a forward algorithm, until convergence to a limit cycle:
	\begin{enumerate}
		\item Compute the marginal $\bar p_t$ given the current solution for $\bar p_{t-1}$ and $\pi_{t-1}$, by applying \eqref{eq:forward}.
		\item Compute the optimal policy $\pi_t$ given the current solution for $\bar p_t$, $\bar\pi$ and $\nu$, by applying \eqref{eq:opt}.
	\end{enumerate}
\end{enumerate}

\section{Periodicity in reactive policies}
\label{sec:periodic}

Throughout the previous sections, we always referred to periodic reactive policies rather than stationary ones, even though the POMDP itself is assumed to be stationary.
Periodic reactive policies may seem to be a contradiction in terms, since their actions depend not only on the most recent observation, but also on the time $t$.
They require a clock to be available to the actuator, with period that is a multiple of the policy period.

We argue that periodic policies must inevitably be a part of the solution concept of POMDPs with selective attention.
When paying full attention to inputs, in the form of exact Bayesian inference, we can restrict the discussion to stationary policies \cite{shani2013survey}.
When attention is partial, there are significant drawbacks to considering only stationary policies.

\begin{figure}
\center
\begin{tikzpicture}
\node (l)		[w]						{};
\node (r)		[w, right of=l, xshift=7em]		{};
\draw (l.20) -- (r.160)		[e] node [midway, above, scale=.7] {right (+reward)};
\draw (r.200) -- (l.340)	[e] node [midway, below, scale=.7] {left (+reward)};
\draw (l) to [loop left, looseness=20] (l)	[e] node [xshift=-2em, yshift=1em, scale=.7] {left};
\draw (r) to [loop right, looseness=20] (r)	[e] node [xshift=2em, yshift=1.55em, scale=.7] {right};
\end{tikzpicture}
\caption{An unobservable POMDP, having an optimal policy of period 2}
\label{fig:periodic}
\end{figure}
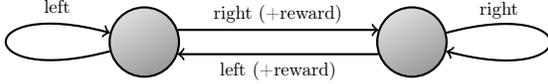

One drawback is that the best stationary policy is generally arbitrarily worse than the optimal periodic policy.
Adapting the example in \cite{singh1994learning}, consider the POMDP illustrated in Figure \ref{fig:periodic}.
This model has 2 states, 1 (uninformative) observation and 2 actions.
The actions deterministically set the next state, and a reward (negative cost) is given for switching to the other state.

The optimal stationary retentive policy for this POMDP is to have two internal memory states, each indicating a different action, and switch between them in each time step.
This policy gets the reward in each time step, but incurs 1 bit of internal cost\footnote{See \cite{fox2012bounded} for the definition of the internal cost of a retentive policy.}.

On the other hand, a stationary reactive policy in an unobservable POMDP is just a fixed distribution over the actions, and it can be no better in this instance than the uniform distribution.
This policy yields only half the expected reward, but incurs no internal cost.

Lastly, the reactive policy of period 2 which alternates between the actions also receives the full reward, at seemingly no internal cost.
In fact, this would seemingly also be the preferred retentive solution, if the internal cost is taken into consideration.

Of course, counting no internal cost for a periodic policy is cheating.
Instead of paying attention to its sensors or memory, the agent is paying attention to a clock, but that attention is still a burden on internal resources.

Similar to the informational cost between $o_t$ and $a_t$, we need to add a term for the informational cost between $t$ and $a_t$.
For a reactive policy with period $\C T$, this cost term can naturally be defined by
$$\I[t;a_t]=\frac1{\C T}\sum_{t=0}^{\C T-1}\DKL[\bar\pi_t\|\bar\pi]$$
$$=\frac1{\C T}\sum_{t=0}^{\C T-1}\sum_{a_t}\bar\pi_t(a_t)\log\frac{\bar\pi_t(a_t)}{\bar\pi(a_t)},$$
with
$$\bar\pi(a)=\frac1{\C T}\sum_{t=0}^{\C T-1}\bar\pi_t(a).$$
Here we use the fact that the phase of the cycle is distributed uniformly during the process.

Adding the term $\I[t;a_t]$ to the free energy is equivalent to asserting that a clock is observable to the agent, and that attention to it is as costly as to any other part of the observation.
The pointwise informational cost is now
$$\tilde i_t(o_t,a_t)=\log\frac{\pi_t(a_t|o_t)}{\bar\pi(a_t)},$$
and the average expected internal cost is
$$\tilde{\C I}=\frac1{\C T}\sum_{t=0}^{\C T-1}\I[o_t;a_t]+\I[t;a_t]$$
$$=\I[o_t;a_t|t]+\I[t;a_t]=\I[t,o_t;a_t].$$
The values of $\tilde f_t$, $\tilde\nu_t$ and $\tilde d_t$ change accordingly, and the optimal policy is now
$$\pi_t(a_t|o_t)=\frac1{\tilde Z_t(o_t)}\bar\pi(a_t)\exp(-\beta\tilde d_t(o_t,a_t)),$$
with the proper partition function $\tilde Z_t(o_t)$.

\begin{figure}
\includegraphics[width=\columnwidth]{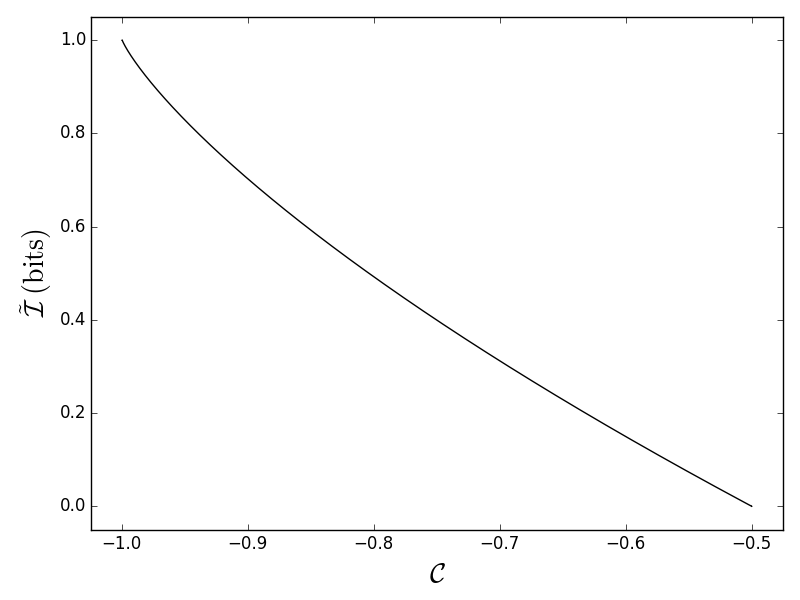}
\caption{Information-cost curve for the POMDP in Figure \ref{fig:periodic}; Points on the curve were achieved by the algorithm with different values of $\beta$, and points above it are achievable; The curve is convex, with slope $-\beta$}
\label{fig:periodic-rd}
\end{figure}

\begin{figure}
\includegraphics[width=\columnwidth]{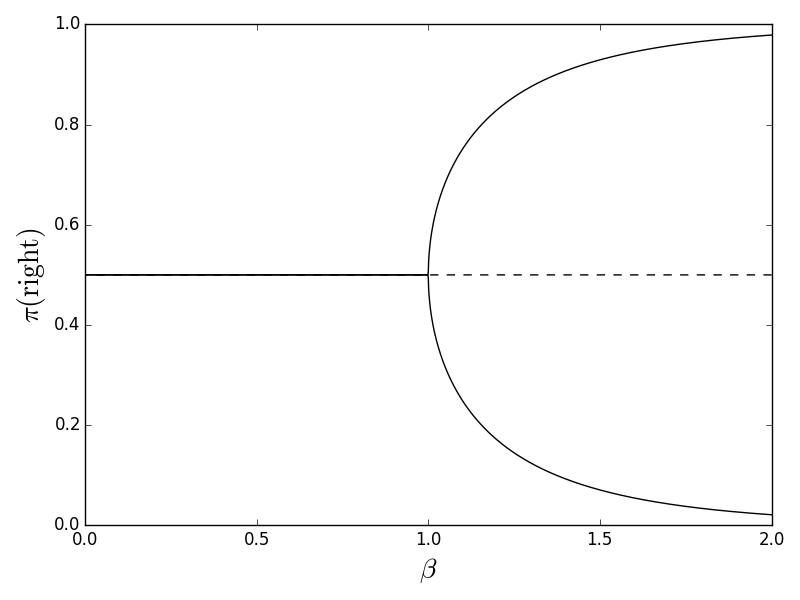}
\caption{Finite-state diagram for the iterative algorithm applied to the POMDP in Figure \ref{fig:periodic}, as a function of the cost conversion rate $\beta$; Points on the curve are the probability of taking the action "right" in each phase of the limit cycle of the algorithm, when run to convergence with the given $\beta$}
\label{fig:periodic-res}
\end{figure}

This allows us to consider policies which are "softly periodic", in that they attend to some but not all time information.
Figure \ref{fig:periodic-rd} shows the information-cost curve for the POMDP in Figure \ref{fig:periodic}, and Figure \ref{fig:periodic-res} shows the final-state diagram for the iterative algorithm with the schedule in Section \ref{sec:algorithm}.

Interestingly, this problem exhibits a bifurcation at $\beta=1$.
Below this value, information is too costly, and the optimal solution is the stationary uniform policy.
At $\beta=1$, the system undergoes a period-doubling bifurcation, and above this value the optimal policy becomes periodic with period 2 --- the two phases of the cycle are given by the two branches in Figure \ref{fig:periodic-res}.
The "hardness" of this periodicity, as measured by the information $I[t;a_t]$, grows continuously from 0, and tends to 1 bit as $\beta$ tends to infinity.

Above the critical point, a third solution exists, which is a fixed point of the optimization schedule (the dashed line in Figure \ref{fig:periodic-res}).
This solution is the optimal stationary reactive policy, but it is an unstable fixed point: starting the optimization from a small perturbation of this solution does not converge back to it, but diverges until it reaches the periodic solution.
Thus we have a supercritical pitchfork bifurcation \cite{strogatz2001nonlinear}.

The instability of the stationary solution is another paramount reason for allowing periodic policies.
It would be practically impossible to find a stationary solution using Bellman-like variational methods, as the one presented in this paper.
In contrast, approaches such as policy-gradient methods \cite{aberdeen2003policy} generally can find stationary solutions, but these are generally not fixed points of a Bellman recursion, and are thus not Bellman-optimal~\cite{bellman1957dynamic}.

\begin{figure}
\center
\begin{tikzpicture}
\node (ll)		[w, rounded rectangle, minimum height=3em, minimum width=8em]						{\begin{tabular}{c}left end\\loaded\end{tabular}};
\node (rl)		[w, rounded rectangle, right of=ll, xshift=4em, minimum height=3em, minimum width=8em]		{\begin{tabular}{c}right end\\loaded\end{tabular}};
\node (lu)		[w, rounded rectangle, below of=ll, xshift=-2em, yshift=-1.5em, minimum height=3em, minimum width=8em]				{\begin{tabular}{c}left end\\unloaded\end{tabular}};
\node (ru)		[w, rounded rectangle, right of=lu, xshift=4em, minimum height=3em, minimum width=8em]		{\begin{tabular}{c}right end\\unloaded\end{tabular}};
\coordinate	[right of=rl, xshift=1em]		(lsl);
\coordinate	[right of=ru, xshift=1em]		(lsu);
\coordinate	[below of=lu, yshift=1em]		(psl);
\coordinate	[below of=ru, yshift=1em]		(psr);
\draw (ll.5) -- (rl.175)		[e] node [midway, above, scale=.7] {right};
\draw (rl.185) -- (ll.355)	[e] node [midway, below, scale=.7] {left};
\draw (lu.100) -- (ll.260)	[e] node [midway, left, scale=.7] {load};
\draw (ll.280) -- (lu.80)	[e] node [midway, right, scale=.7] {unload};
\draw (rl.270) -- (ru.90)	[e] node [midway, right, scale=.7] {\begin{tabular}{c}unload\\(reward)\end{tabular}};
\draw (lu.5) -- (ru.175)	[e] node [midway, above, scale=.7] {right};
\draw (ru.185) -- (lu.355)	[e] node [midway, below, scale=.7] {left};
\draw (lsl) -- (lsu)		[de] node [midway, xshift=.5em, scale=.7, rotate=90]	{load sensor};
\draw (psl) -- (psr)		[de] node [midway, below, scale=.7] {location sensor};
\end{tikzpicture}
\caption{A POMDP of a robot moving items from the left end of a corridor to the right one; Shown actions succeed with probability 0.8, otherwise the state remains the same; Location sensor correct with probability 0.88, load sensor with probability 0.7}
\label{fig:robot}
\end{figure}
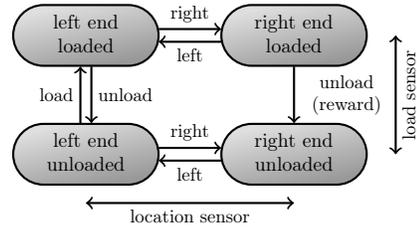

\begin{figure}
\includegraphics[width=\columnwidth]{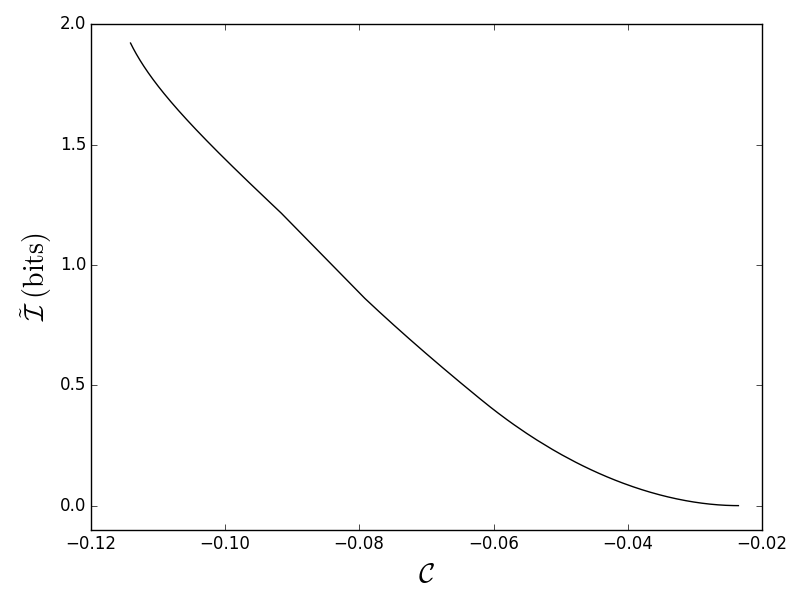}
\caption{Information-cost curve for the POMDP in Figure \ref{fig:robot}}
\label{fig:robot-rd}
\end{figure}

\begin{figure}
\includegraphics[width=\columnwidth]{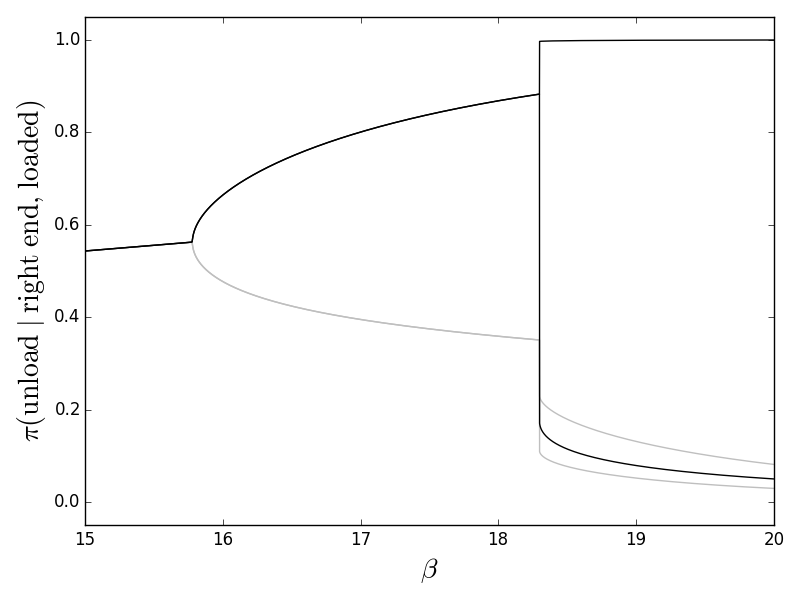}
\caption{Finite-state diagram for the POMDP in Figure \ref{fig:robot}; Points on the curve are the probability of taking the action "unload" when the sensors indicate "right end" and "loaded", in each phase of the limit cycle; The lower branch is grayed out for clarity}
\label{fig:robot-res}
\end{figure}

\subsection{Robot example}

As another example, consider the POMDP illustrated in Figure \ref{fig:robot}.
Here a robot is engaged in moving items from the left end of a corridor to the right one.
The robot can be in one of 4 states: it can be at either end of the corridor, and it can be carrying an item or not.
It has 4 actions: to move to the left end of the corridor, or to the right, or to pick up or put down an item.
However, an action can fail with probability 0.2, leaving the robot at the same state.
The robot can only pick up an item at the left end of the corridor, and it receives a reward for dropping an item at the right end.

The robot has 4 possible observations from two binary sensors, telling it its position and whether or not it is carrying an item.
The location sensor is more reliable, showing the correct position with probability 0.88.
The load sensor only shows the correct load state with probability 0.7.
The parameters were selected for visual clarity of the results.

Figure \ref{fig:robot-rd} shows the information-cost curve for this problem.
Here there are two phase transitions, where the period doubles to 2, then again to 4 (Figure \ref{fig:robot-res}).
When attention is scarce, the robot's actions are more uniformly random.
In this situation, the sensors, although noisy, carry more relevant information than the clock, since they better correspond to the actual state.

As attention increases with $\beta$, the robot relies more and more on its sensors.
Conditional on the observation, the robot makes its actions less and less stochastic.
At some point, the policy is reliable enough that the clock has more relevant information about the load state than the noisy sensor.
At that point, a pitchfork bifurcation occurs, and the robot begins to rely mostly on the parity of the clock to decide when to move, and on the location sensor to decide where to move and whether to load or unload.

With the parameters above, as $\beta$ keeps increasing, the clock eventually becomes even more reliable than the location sensor, and a second period doubling occurs, to period 4.
Asymptotically as $\beta$ grows to infinity, the clock signal takes precedence over both sensors, and the agent unloads its cargo on schedule even if its sensors tells it that it is dislocated or empty handed.

\section{Discussion}

In this paper we presented three novel results involving reactive agents interacting with partially observable systems.
We have motivated the focus on reactive policies through a reduction from retentive policies, introduced a principle and an algorithm for optimizing reactive policies, and explored a surprising aspect of their phenomenology.

We conclude with a few remarks on the implication of each contribution.

\subsection{Selective attention as clustering}

Information-constrained clustering can also be construed as source coding \cite{rose1998deterministic}, so that the data to be clustered is considered the source, and the cluster centroids the reconstruction.
Following the relation we show between selective attention and source coding, we can think of a reactive policy as a soft clustering of observations into actions.

With the information constraint removed, the clustering becomes hard, mapping each data point to its closest centroid.
Similarly in our case, as $\beta$ grows the policy becomes more deterministic, until at $\beta\to\infty$ it always picks the optimal action for each observation.

The implication of viewing reactive policies as clustering is that actions should generally be simpler, and never more complex, than the observations on which they are based.
Indeed, there is a duality between observations and actions, and between selective attention (the retained part of the observation) and selective action (the intended part of the action, as divergence from the prior $\bar\pi$).
Information that is not retained cannot be used for choosing actions, and there is no point in retaining information that is not used.

\subsection{Implications of selective attention for retentive agents}

In this paper we have focused on reactive agents, and introduced the minimum-information principle for optimal selective attention.
However, as the reduction in Section \ref{sec:reduction} shows, this has implications for retentive agents as well.

The effect of selective attention is to make internal states less complex than their inputs, by discarding information that is not useful enough.
When applied to the inference policy, this leads to approximate inference, that trades off the external value of information in guiding actions with its internal cost in information complexity.
In fact, an inference process in POMDPs is equivalent to sequential clustering.
With each new observation $o_t$, the pair $(m_{t-1},o_t)$ is clustered into a new internal state $m_t$.

The major challenge when planning in POMDPs is approximating the Bayesian belief in such a way that allows efficient planning and execution, while not losing too much value.
Selective attention, and in this case retention, is precisely such a principle.
The application of this approach to retentive agents is left for future work.

\subsection{Policy bifurcations and chaos theory}

We have discovered the occurrence of bifurcations in the optimization process of reactive policies.
It presents many of the characterizing features of chaos theory of iterated functions, such as period doubling and slow convergence near the bifurcation points.
We expect to see many more such features in other, more complex systems.
We conjecture that systems with more states, perhaps infinitely many, can present a cascade of bifurcations, leading to aperiodicity and chaos.

A full investigation of the bearings of the theories of bifurcation and chaos to optimal control in dynamical systems is beyond the scope of this report.
To the extent that such a connection exists, it could be of profound philosophical implications, as it could indicate that intelligent agents interacting with complex environments must choose among the following alternatives:
\begin{itemize}
	\item Plan with very little attention of their inputs
	\item Plan for very short horizons
	\item Plan with some degree of inability to identify their own value function or predict their own future actions.
\end{itemize}


\clearpage
\small
\bibliography{resources}{}

\begin{thebibliography}{10}

\bibitem{aberdeen2003revised}
Douglas Aberdeen.
\newblock A (revised) survey of approximate methods for solving partially
  observable {M}arkov decision processes.
\newblock {\em National ICT Australia, Canberra, Australia}, 2003.

\bibitem{aberdeen2003policy}
Douglas~A. Aberdeen.
\newblock {\em Policy-gradient algorithms for partially observable Markov
  decision processes}.
\newblock PhD thesis, Australian National University, 2003.

\bibitem{bellman1957dynamic}
Richard Bellman.
\newblock {\em Dynamic programming}.
\newblock Princeton University Press, 1957.

\bibitem{boyd2004convex}
Stephen Boyd and Lieven Vandenberghe.
\newblock {\em Convex optimization}.
\newblock Cambridge University Press, 2004.

\bibitem{cover2012elements}
Thomas~M. Cover and Joy~A. Thomas.
\newblock {\em Elements of information theory}.
\newblock Wiley Series in Telecommunications and Signal Processing. John Wiley
  \& Sons, 2012.

\bibitem{fox2012bounded}
Roy Fox and Naftali Tishby.
\newblock Bounded planning in passive {POMDP}s.
\newblock In {\em Proceedings of the International Conference on Machine
  Learning (ICML)}, pages 1775--1782, 2012.

\bibitem{jaakkola1995reinforcement}
Tommi Jaakkola, Satinder~P. Singh, and Michael~I. Jordan.
\newblock Reinforcement learning algorithm for partially observable {M}arkov
  decision problems.
\newblock In {\em Proceedings of the Advances in Neural Information Processing
  Systems (NIPS)}, pages 345--352, 1995.

\bibitem{jaynes2003probability}
Edwin~T. Jaynes.
\newblock {\em Probability theory: The logic of science}.
\newblock Cambridge University Press, 2003.

\bibitem{kappen2012optimal}
Hilbert~J. Kappen, Vicen{\c{c}} G{\'o}mez, and Manfred Opper.
\newblock Optimal control as a graphical model inference problem.
\newblock {\em Machine Learning}, 87(2):159--182, 2012.

\bibitem{littman1994memoryless}
Michael~L. Littman.
\newblock Memoryless policies: Theoretical limitations and practical results.
\newblock In {\em From Animals to Animats 3: Proceedings of the third
  international conference on simulation of adaptive behavior (SAB)}, volume~3,
  page 238, 1994.

\bibitem{meuleau1999solving}
Nicolas Meuleau, Kee-Eung Kim, Leslie~Pack Kaelbling, and Anthony~R Cassandra.
\newblock Solving pomdps by searching the space of finite policies.
\newblock In {\em Proceedings of the Fifteenth conference on Uncertainty in
  artificial intelligence}, pages 417--426. Morgan Kaufmann Publishers Inc.,
  1999.

\bibitem{murphy2000survey}
Kevin~P. Murphy.
\newblock A survey of {POMDP} solution techniques.
\newblock 2000.

\bibitem{osullivan1998alternating}
Joseph~A. O'Sullivan.
\newblock Alternating minimization algorithms: from {B}lahut-{A}rimoto to
  expectation-maximization.
\newblock In {\em Codes, Curves, and Signals}, pages 173--192. Springer, 1998.

\bibitem{pineau2003point}
Joelle Pineau, Geoffrey Gordon, and Sebastian Thrun.
\newblock Point-based value iteration: an anytime algorithm for {POMDP}s.
\newblock In {\em Proceedings of the International Joint Conference on
  Artificial Intelligence (IJCAI)}, pages 1025--1032, 2003.

\bibitem{poupart2003bounded}
Pascal Poupart and Craig Boutilier.
\newblock Bounded finite state controllers.
\newblock In {\em Advances in neural information processing systems}, page
  None, 2003.

\bibitem{rose1998deterministic}
Kenneth Rose.
\newblock Deterministic annealing for clustering, compression, classification,
  regression, and related optimization problems.
\newblock {\em Proceedings of the IEEE}, 86(11):2210--2239, 1998.

\bibitem{roy2005finding}
Nicholas Roy, Geoffrey~J. Gordon, and Sebastian Thrun.
\newblock Finding approximate {POMDP} solutions through belief compression.
\newblock {\em Journal of Artificial Intelligence Research (JAIR)}, 23:1--40,
  2005.

\bibitem{rubin2012trading}
Jonathan Rubin, Ohad Shamir, and Naftali Tishby.
\newblock Trading value and information in {MDP}s.
\newblock In {\em Decision Making with Imperfect Decision Makers}, pages
  57--74, 2011.

\bibitem{shani2013survey}
Guy Shani, Joelle Pineau, and Robert Kaplow.
\newblock A survey of point-based {POMDP} solvers.
\newblock {\em Autonomous Agents and Multi-Agent Systems (AAMAS)}, 27(1):1--51,
  2013.

\bibitem{singh1994learning}
Satinder~P. Singh, Tommi Jaakkola, and Michael~I. Jordan.
\newblock Learning without state-estimation in partially observable {M}arkovian
  decision processes.
\newblock In {\em Proceedings of the International Conference on Machine
  Learning (ICML)}, pages 284--292, 1994.

\bibitem{strogatz2001nonlinear}
Steven~H. Strogatz.
\newblock {\em Nonlinear dynamics and chaos: With applications to physics,
  biology, chemistry, and engineering}.
\newblock Perseus publishing, 2001.

\bibitem{sutton1998introduction}
Richard~S. Sutton and Andrew~G. Barto.
\newblock {\em Introduction to reinforcement learning}.
\newblock MIT Press, 1998.

\bibitem{tatikonda1998control}
Sekhar Tatikonda, Anant Sahai, and Sanjoy Mitter.
\newblock Control of {LQG} systems under communication constraints.
\newblock In {\em Proceedings of the IEEE Conference on Decision and Control},
  volume~1, pages 1165--1170, 1998.

\bibitem{tishby2011information}
Naftali Tishby and Daniel Polani.
\newblock Information theory of decisions and actions.
\newblock In {\em Perception-Action Cycle}, pages 601--636. Springer, 2011.

\bibitem{todorov2006linearly}
Emanuel Todorov.
\newblock Linearly-solvable {M}arkov decision problems.
\newblock In {\em Proceedings of the Advances in Neural Information Processing
  Systems (NIPS)}, pages 1369--1376, 2006.

\end{thebibliography}
\bibliographystyle{plain}

\end{document}